\def\year{2021}\relax
\documentclass[letterpaper]{article}
\usepackage{aaai21} 
\usepackage{times} 
\usepackage{helvet} 
\usepackage{courier} 
\usepackage[hyphens]{url} 
\usepackage{graphicx} 
\usepackage{graphicx}  
\usepackage{natbib}
\usepackage{caption}
\usepackage[paper,cmrgreekup]{pdef}
\usepackage[ruled,vlined]{algorithm2e}
\usepackage{booktabs}       
\usepackage{caption}
\usepackage{subcaption}
\usepackage[normalem]{ulem}
\usepackage{xcolor}
\usepackage{multirow}
\usepackage{makecell}
\usepackage{bm}
\usepackage{tikz}
\usepackage{xparse}

\title{Feature Space Singularity for Out-of-Distribution Detection}
\author{Haiwen Huang\textsuperscript{\rm 1}, Zhihan Li\textsuperscript{\rm 2}, Lulu Wang\textsuperscript{\rm 3}, Sishuo Chen\textsuperscript{\rm 2}\\
Bin Dong\textsuperscript{\rm 2, 4, 5}, Xinyu Zhou\textsuperscript{\rm 3}\\
}
\affiliations{
    \textsuperscript{\rm 1} Department of Computer Science, University of Oxford\\
    \textsuperscript{\rm 2} Peking University \hspace{1cm}\textsuperscript{\rm 3} MEGVII Technology\\
     \textsuperscript{\rm 4}Beijing International Center for Mathematical Research\\
     \textsuperscript{\rm 5}Institute for Artificial Intelligence and Center for Data Science\\

    haiwen.huang2@cs.ox.ac.uk,  zxy@megvii.com, dongbin@math.pku.edu.cn

}

\makeatletter
\def\sim@y@offset{.025}
\def\sim@x@scale{.35}
\def\sim@y@scale{.125}
\def\sim@y@thick{.008}

\newsavebox\sim@upper
\newsavebox\sim@lower

\NewDocumentCommand{\xSim}{ O{} m }{%
  \TextOrMath{%
    \PackageError{TEST}{`\string\xSim` is valid in math mode only.}{}%
  }{
    \sbox\sim@upper{\scalebox{0.7}{\hbox{$#2$}}}
    \sbox\sim@lower{\scalebox{0.7}{\hbox{$#1$}}}
    \pgfmathparse{min(max(\wd\sim@upper/1em, \wd\sim@lower/1em, 1.0), 1.5)}
    \edef\sim@ratio{\pgfmathresult}
    \def\sim@x {\sim@x@scale * \sim@ratio}
    \def\sim@@@y {\sim@y@offset * \sim@ratio}
    \def\sim@@@@y {-\sim@y@offset * \sim@ratio}
    \def\sim@y {\sim@y@scale * \sim@ratio}
    \def\sim@@y{\sim@y@thick * \sim@ratio}
    \pgfmathparse{floor(max(\wd\sim@upper/1em, \wd\sim@lower/1em)) - 0.5}
    \edef\sim@wd{\pgfmathresult em}
    \mathrel{
      \begin{tikzpicture}[baseline=-.7ex]
        \filldraw[line width=.2pt] 
          (0, \sim@@@y)
            .. controls +(\sim@x, \sim@y+\sim@@y) and +(-\sim@x, -\sim@y) .. 
          +(\sim@wd, 0) 
            node[midway, above] {\usebox\sim@upper}
            .. controls +(-\sim@x, -\sim@y-\sim@@y) and +(\sim@x, \sim@y) .. 
          (0, \sim@@@y);
          \filldraw[line width=.2pt] 
            (0, \sim@@@@y)
              .. controls +(\sim@x, \sim@y+\sim@@y) and +(-\sim@x, -\sim@y) .. 
            +(\sim@wd, 0)  
              node[midway, below] {\usebox\sim@lower}
              .. controls +(-\sim@x, -\sim@y-\sim@@y) and +(\sim@x, \sim@y) .. 
            (0, \sim@@@@y);
      \end{tikzpicture}
    }
  }%
}
\makeatother

\begin{document}

\maketitle

\begin{abstract}
Out-of-Distribution (OoD) detection is important for building safe artificial intelligence systems.
However, current OoD detection methods still cannot meet the performance requirements for practical deployment.
In this paper, we propose a simple yet effective algorithm based on a novel observation: in a trained neural network, OoD samples with bounded norms well concentrate in the feature space.
We call the center of OoD features the \emph{Feature Space Singularity (FSS)}, and denote the distance of a sample feature to FSS as \emph{FSSD}.
Then, OoD samples can be identified by taking a threshold on the FSSD.
Our analysis of the phenomenon  reveals why our algorithm works.
We demonstrate that our algorithm achieves state-of-the-art performance on various OoD detection benchmarks. Besides, FSSD also enjoys robustness to slight corruption in test data and can be further enhanced by ensembling. These make FSSD a promising algorithm to be employed in real world. We release our code at \url{https://github.com/megvii-research/FSSD_OoD_Detection}.

\end{abstract}

\section{Introduction}\label{introduction}

Empirical risk minimization fits a statistical model on a training set which is independently sampled from the  data distribution.
As a result, the yielded model is expected to generalize to in-distribution data drawn from the same distribution.
However, in real applications, it is inevitable for a model to make predictions on \emph{Out-of-Distribution (OoD)} data instead of in-distribution data on which the model is trained.
This can lead to fatal errors such as over-confident or ridiculous
predictions ~\cite{whyRelu, FailingLoudly}.
Therefore, it is crucial to understand the uncertainty of models and automatically detect OoD data.
In applications like autonomous driving and medical services, if the model knows what it does not know, human intervention can be sought and security can be significantly improved.

Consider one particular example of OoD detection: some high-quality human face images are given as in-distribution data (\emph{training set} for OoD detector), and we are interested in filtering out non-faces and low quality faces from a large pool of data in the wild (\emph{test set}) in order to ensure reliable prediction.
One natural solution is to remove test samples far from the training data in some designated distances~\cite{mahalanobis, duq}.
However, calculating the distance to the whole training set needs a formidable amount of computation without some special design in feature and architecture, e.g., training a RBF network~\cite{duq}.
\begin{figure*}[t]
\begin{subfigure}{0.65\textwidth}
  \centering
  \includegraphics[width=1\linewidth]{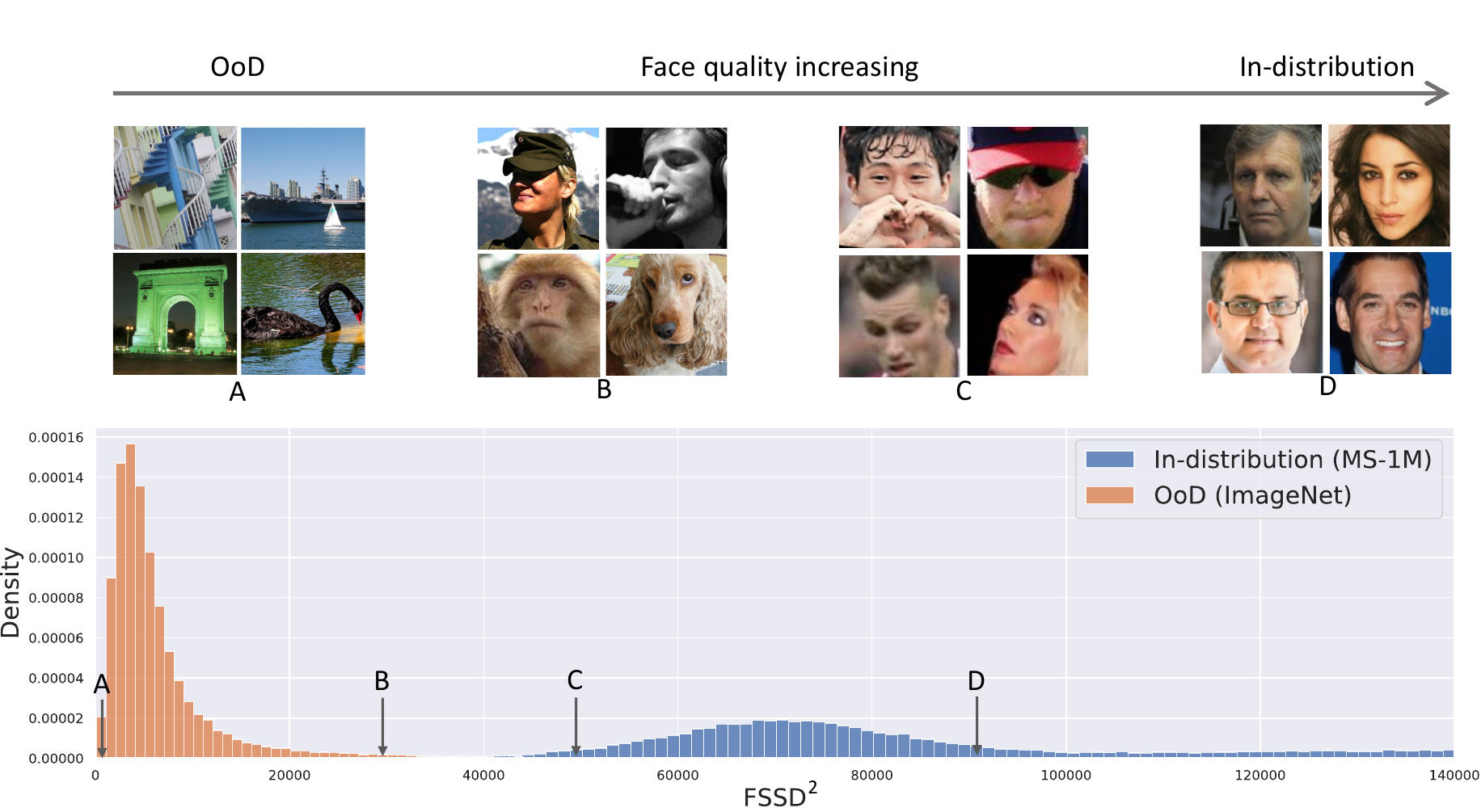}  
\end{subfigure}
\begin{subfigure}{.32\textwidth}
  \centering
  \includegraphics[width=1\linewidth]{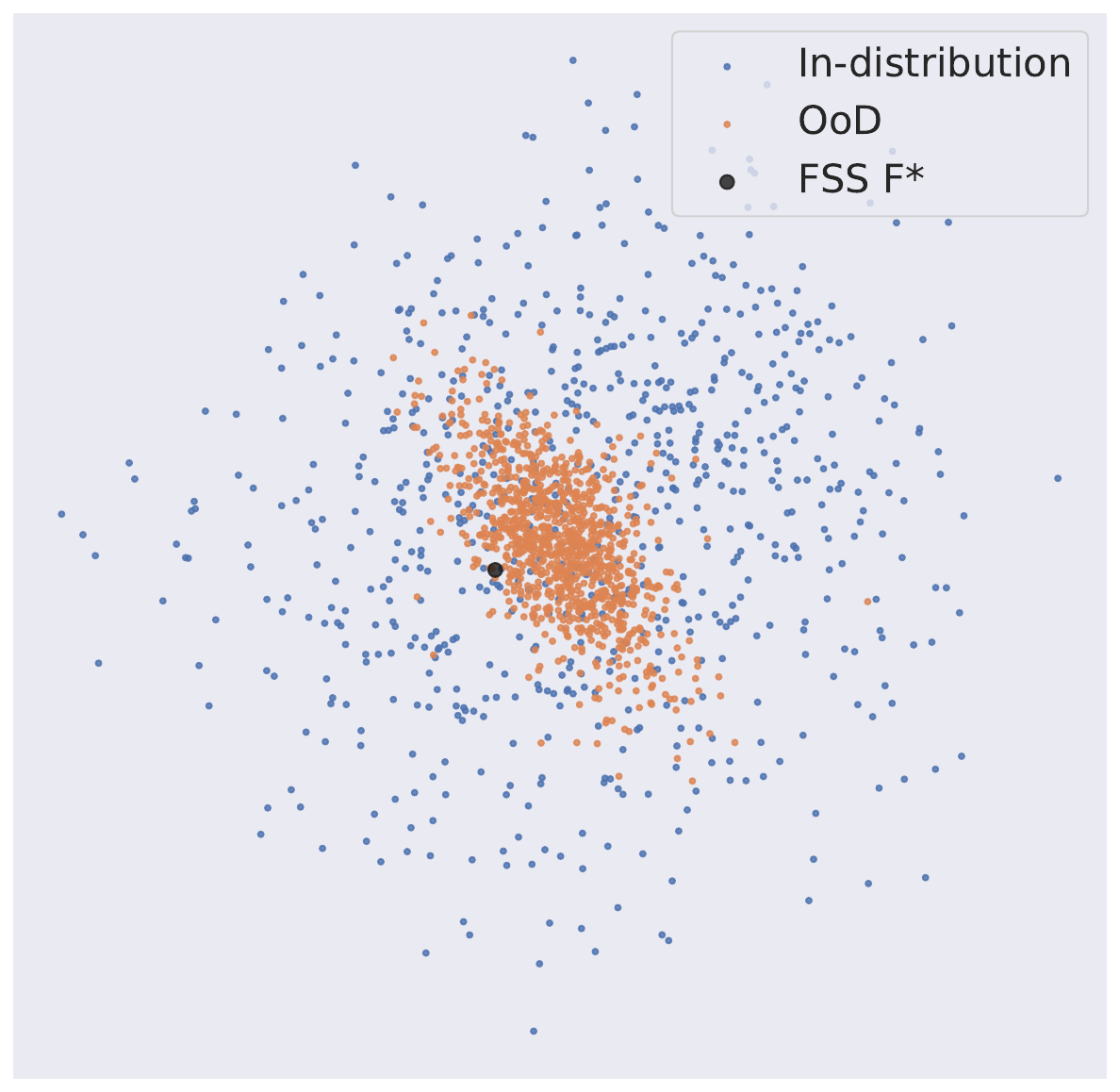}

\end{subfigure}
\caption{Left: Histogram of FSS Distance (FSSD) of MS1M (in-distribution) and ImageNet (OoD). Exemplar images are shown at different FSSDs. We can see that FSSD reflects the OoD degree: as the FSSD increases, images change from non-faces and  pseudo-faces, to low-quality faces and high-quality faces. Right: Principle components of features from the penultimate layer. The spatial relationship among FSS, OoD data, and in-distribution data is shown.}
\label{fig:phenomena}
\end{figure*}
In this paper, we present a simple yet effective distance-based solution, which neither computes the distance to the training data nor performs extra model training than a standard classifier.

Our approach is based on a novel observation about OoD samples:
\begin{quote}
\emph{In a trained neural network, OoD samples with bounded norms well concentrate in the feature space of the neural network.}
\end{quote}
In Figure~\ref{fig:phenomena}, we show an example where OoD features from ImageNet~\cite{imagenet} concentrate in a neural network trained on the facial dataset MS-1M~\cite{ms1m}. Figure 2 and 3 provide more examples of this phenomenon. In fact, we find this phenomenon to be universal in most training configurations for most datasets.

To be more precise, for a given  feature extractor $F_\theta$ trained on in-distribution data, the observation states that there exists a point $F^\ast$ in the output space
of $F_\theta$ such that $ \norm{ F_{\theta} \rbr{x} - F^{\ast} } $ is small for $ x \in \mathcal{X}_{\text{OoD}}$, where $\mathcal{X}_{\text{OoD}}$ is the set of OoD samples.
We call $F^\ast$ the \emph{Feature Space Singularity (FSS)}.
Moreover, we discover  the \emph{FSS Distance (FSSD)} 
\begin{equation}
\text{FSSD} \rbr{x} := \norm{ F_{\theta} \rbr{x} - F^{\ast} }
\label{fssd}
\end{equation}
can reflect the degree of OoD, and thus can be readily used as a metric for OoD detection.

Our analysis demonstrates that this phenomenon can be explained by the training dynamics. The key observation is that FSSD can be seen as an approximate movement of $F_{\theta_t} \rbr{x}$ during training, where $F^*$ is the initial concentration point of the features. The difference in the moving speed $\frac{ \sd F_{\theta_t} \rbr{x} }{ \sd t }$ stems from the different similarity to the training data measured by the inner product of the gradients. Moreover, this similarity measure varies according to the architecture of the feature extractor. 

We demonstrate the effectiveness of our proposed method with multiple neural networks (LeNet~\cite{mnist}, ResNet~\cite{resnet}, ResNeXt~\cite{resnext}, LSTM~\cite{lstm}) trained on various datasets for classification (FashionMNIST~\cite{fmnist}, CIFAR10~\cite{cifar10}, ImageNet~\cite{imagenet}, CelebA~\cite{celeba}, MS-1M~\cite{ms1m}, bacteria genome dataset~\cite{ren2019likelihood}) with varying training set sizes.
We show that FSSD achieves state-of-the-art performance on almost all the considered benchmarks.
Moreover, the performance margin between FSSD and other methods increases as the size of the training set increases.
In particular, on large-scale benchmarks (CelebA and MS-1M), FSSD advances the AUROC by about 5\%.
We also evaluate the robustness of our algorithm when test images are corrupted.
We find that our algorithm can still reliably detect OoD samples under this circumstance.
Finally, we investigate the effects of ensembling FSSDs from different layers of a single neural network and multiple trained netowrks.

We summarize our contributions as follows.
\begin{itemize}
\item We observe that in feature spaces of trained networks OoD samples concentrate near a point (FSS), and the distance from a sample feature to FSS (FSSD) measures the degree of OoD (Section 1).
\item We analyze the concentration phenomenon by analyzing the dynamics of in-distribution and OoD features during training (Section 2).
\item We introduce the FSSD algorithm (Section 3) which achieves state-of-the-art performance on various OoD detection benchmarks with considerable robustness (Section 4).
\end{itemize}

\begin{figure*}[t]
    \centering
    \begin{subfigure}{\textwidth}
    	\centering
         \includegraphics[width=1\linewidth]{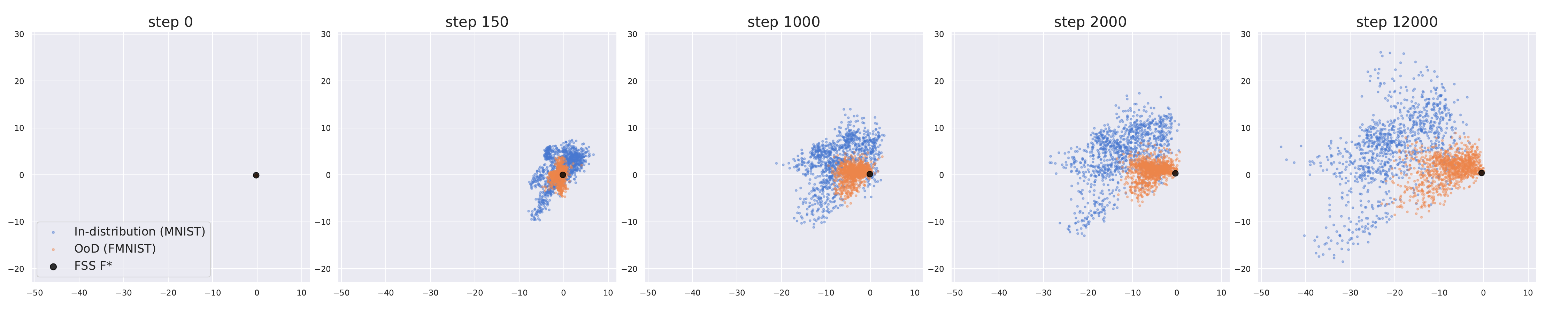}
          \caption{The dynamics of features $F_{\theta_t} \rbr{x} $.}
    \end{subfigure}
\begin{subfigure}{\textwidth}
    \centering
    \includegraphics[width=1\linewidth]{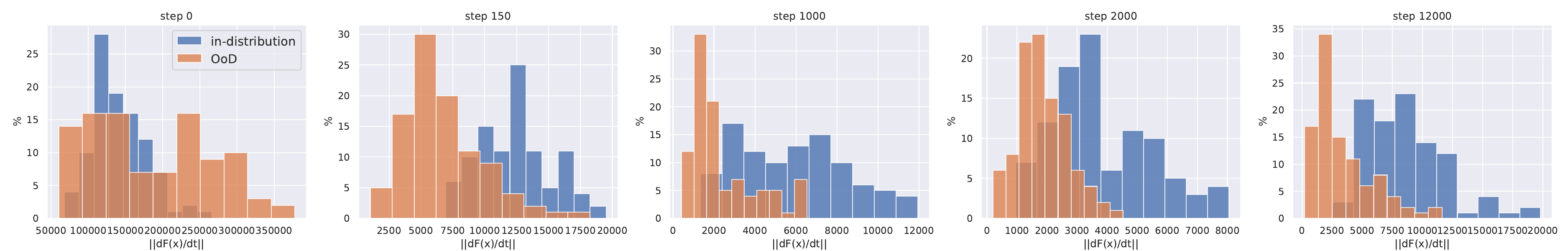}
    \caption{The norm of the derivative, i.e., "moving speed", of last-layer feature vector  at different time steps. }
    \end{subfigure}
     \caption{
     We show first two principle components of the feature vector and the L2 norm of the derivatives (Equation~\eqref{eq:DynFeat}). 
     Features and derivatives are calculated  from the last fully-connected layer of a LeNet trained on MNIST (in-distribution).
     We feed in FashionMNIST data as OoD samples.  
    At initialization, features of both in-distribution and OoD samples concentrate near FSS $F^{\ast}$. After training, features of in-distribution samples are pulled away from FSS $F^{\ast}$, while features of OoD samples remain close to FSS $F^{\ast}$. Similar dynamics of the softmax layer on in-distribution data was observed by \cite{grandtour}.
    }
\end{figure*}

\section{Analyzing and Understanding the Concentration Phenomenon}

In this section, we analyze the concentration phenomenon. The key observation is that during training, the features of the training data are supervised to move away from the initial point, and \textit{the moving speeds of features of other data depend on their similarity to the training data}. Specifically, this similarity is measured by the inner product of the gradients. Therefore, data that are dissimilar to the training data will move little and concentrate in the feature space. This is how FSSD identifies OoD data.

To see this,  we derive the training dynamics of the feature vectors.
We denote $ F_{\theta} : \mathbb{R}^a \rightarrow \mathbb{R}^b $ as the feature extractor which maps inputs to features  and $ G_{\phi} : \mathbb{R}^b \rightarrow \mathbb{R}^c $ to be the map from features to outputs. 
The two mappings are parameterized by $\theta$ and $\phi$ respectively. 
The corresponding loss function can be denoted as $\mathcal{L}_{\phi} \rbr{ F_{\theta} \rbr{x_1}, \cdots, F_{\theta} \rbr{x_M} }$. A popular choice is $ \mathcal{L}_{\phi} \rbr{ F_{\theta} \rbr{x_1}, \cdots, F_{\theta} \rbr{x_M} } = \sum_{ m = 1 }^{M} \ell \rbr{ G_{\phi} \rbr{ F_{\theta} \rbr{x_m} }, y_m } / M $, where $\ell$ is the cross entropy loss or the mean squared error. 
Then, 
the \emph{gradient descent} dynamics of $\theta$ is 
\begin{equation}
\begin{aligned}
\frac{ \sd \theta_t }{ \sd t } =& -\frac{ \pd \mathcal{L}_{\phi} }{ \pd \theta_t } \rbr{ F_{\theta_t} \rbr{x_1}, \cdots, F_{\theta_t} \rbr{x_M} } \\
=& -\sum_{ m = 1 }^M \frac{ \pd F_{\theta_t} \rbr{x_m}^{\mathsf{T}} }{ \pd \theta_t } \pdl{m} \mathcal{L}_{\phi},
\end{aligned}
\end{equation}
where $ \pdl{m} \mathcal{L}_{\phi} = \frac{ \pd \mathcal{L}_{\phi} }{ \pd F_{\theta_t} \rbr{x_m} } \in \mathbb{R}^b $ is the backward propagation gradient and subscript $t$ is the training time.
The dynamics of the feature extractor $F_{\theta}$ as a function is therefore
\begin{equation} \label{eq:DynFeat}
\begin{aligned}
\frac{ \sd F_{\theta_t} \rbr{x} }{ \sd t } = &\frac{ \pd F_{\theta_t} \rbr{x} }{ \pd \theta_t } \frac{ \sd \theta_t }{ \sd t } \\
= &-\sum_{ m = 1 }^{M} \frac{ \pd F_{\theta_t} \rbr{x} }{ \pd \theta_t } \frac{ \pd F_{\theta_t} \rbr{x_m}^{\mathsf{T}} }{ \pd \theta_t } \pdl{m} \mathcal{L}_{\phi}.
\end{aligned}
\end{equation}

From Equation~\eqref{eq:DynFeat}, we can see that the relative moving speed of the feature $F_{\theta_t} \rbr{x}$ depends on the inner product of the \emph{gradient on parameters} between $x$ and the training data $x_m$. Note here $\pdl{m} \mathcal{L}_{\phi}$ is the same for all $x$. Since FSSD defined in Equation~\ref{fssd} can be seen as the integration of $\frac{ \sd F_{\theta_t} \rbr{x} }{ \sd t }$ when the initial value $F_{\theta_0} \rbr{x}$ is $F^*$ for all x, FSSD(x) will also be small when the derivative, i.e., the moving speed, is small.

\begin{figure}
    \centering
    \begin{subfigure}{0.45\textwidth}
         \centering
         \includegraphics[width=\linewidth]{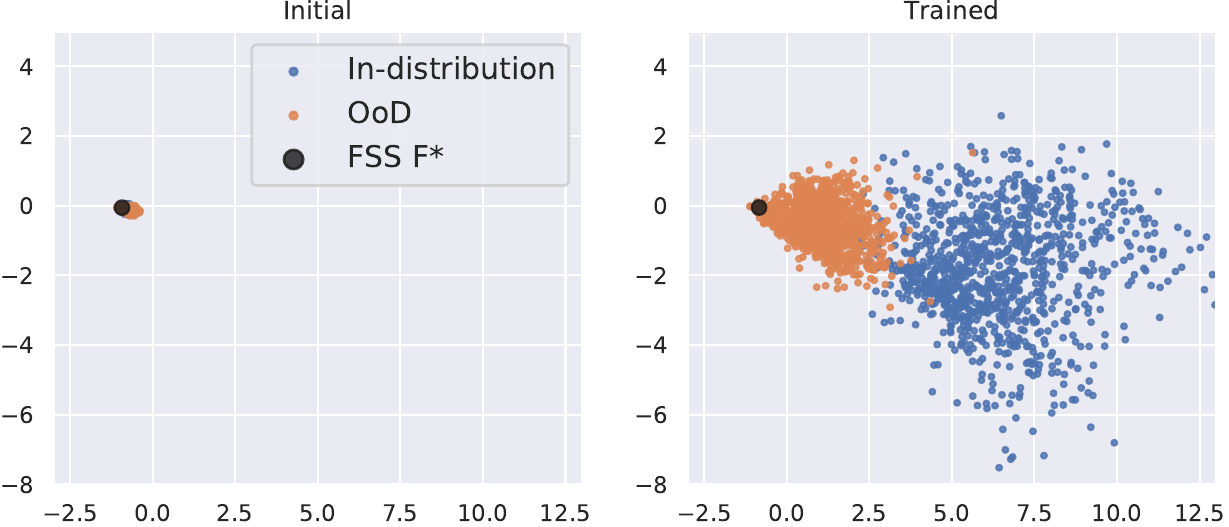}
          \caption{MNIST vs.\ FMNIST}
         \label{concentrate1}
    \end{subfigure}
    \begin{subfigure}{0.45\textwidth}
         \centering
         \includegraphics[width=\linewidth]{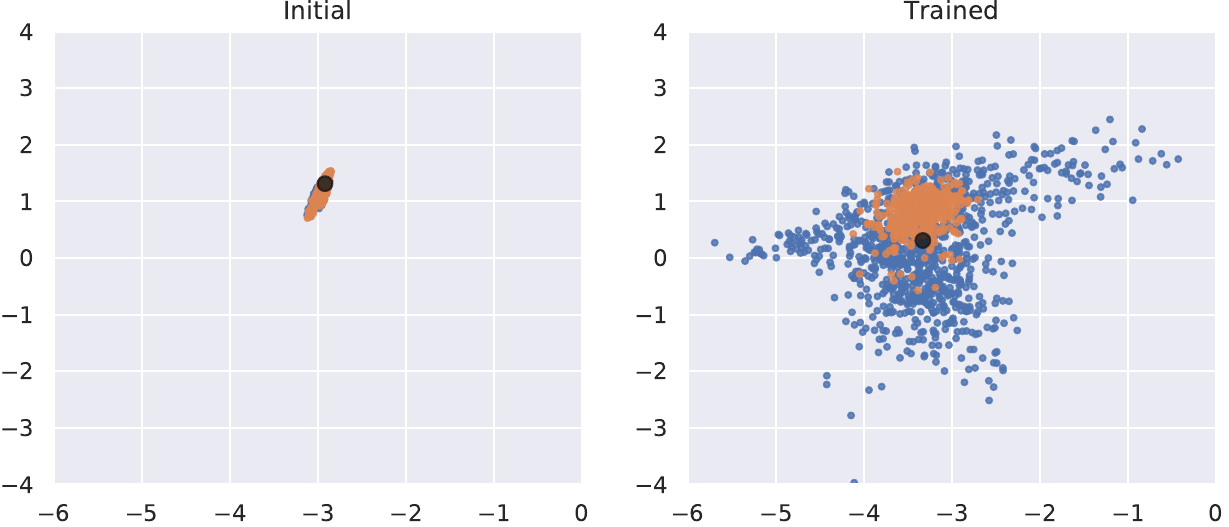}
          \caption{CIFAR10 vs.\ SVHN}
         \label{concentrate3}
    \end{subfigure}
    \caption{Both in-distribution and OoD samples are clustered in the feature space of $ F_{\theta_0} \rbr{x}$ at initialization. Moreover, $ F^{\ast} \approx  F_{\theta_0} \rbr{x}  $ for $ x \in \mathcal{X}_{\text{OoD}} \cup \mathcal{X}_{\text{in-dist}} $.
    }
    \label{concentration}
\end{figure}

In Figure 2, we show both the  features and their moving speeds of in-distribution and OoD data at different steps during training. We can see that although in-distribution and OoD data are indistinguishable at step 0, they are quickly separated since the moving speeds of in-distribution data are larger than those of OoD data (Figure 2(b)) and thus the accumulated movements of in-distribution data are also larger than those of OoD data (Figure 2(a)).
In Figure~\ref{concentration}, we show examples of the initial concentration of features in LeNet and ResNet-34 for MNIST vs. FashionMNIST and CIFAR10 vs. SVHN dataset pairs respectively. Empirically, we find the concentration of both in-distribution and OoD features at the initial stage to be the common case for most popular architectures using random initialization. We show more examples on our Github page.

As we've mentioned, Equation~\eqref{eq:DynFeat} demonstrates that the difference in the moving speed of $F_{\theta_t} \rbr{x}$  stems from difference in  $\Theta_t \rbr{ x, x_m } :=\frac{ \pd F_{\theta_t} \rbr{x} }{ \pd \theta_t } \frac{ \pd F_{\theta_t} \rbr{x_m}^{\mathsf{T}} }{ \pd \theta_t } $. We want to further point out that $\Theta_t \rbr{ x, x_m }$ is effectively acting as a kernel that measures the similarity between $x$ and $x_m$. In fact, when the network width is infinite, $\Theta_t \rbr{ x, x_m }$ will converge to a  time-independent term $\Theta(x, x_m)$, which is called neural tangent kernel (NTK)~\cite{NTK, ntk1, ntk2}. 
In this way, FSSD can be seen as a kernel regression result: 


\begin{equation} \label{eq:FSSDNNApprox}
\begin{aligned}
\text{FSSD} \rbr{x} & \xSim[\text{Equation~\eqref{fssd}}] { F^{\ast} \! \approx \!  F_{\theta_0} \rbr{x} }
\bnorm{\big}{  F_{\theta_t} \rbr{x} - F_{\theta_0} \rbr{x} } \\
& 
= \norm{ \sum_{ m = 1 }^M \int_0^T \Theta_t \rbr{ x, x_m }\pdl{m} \mathcal{L}_{\phi} \sd t } \\
& \approx \norm{ \sum_{ m = 1 }^M\Theta \rbr{ x, x_m } \nu_m},
\end{aligned}
\end{equation} 
where $\nu_m = \int_0^T \pdl{m} \mathcal{L}_{\phi} \sd t$. 

This indicates that the similarity described by the inner product $\Theta_t \rbr{ x, x_m } :=\frac{ \pd F_{\theta_t} \rbr{x} }{ \pd \theta_t } \frac{ \pd F_{\theta_t} \rbr{x_m}^{\mathsf{T}} }{ \pd \theta_t } $ might enjoy similar properties to commonly used kernels such as RBF kernel, which diminishes as the distance between $x$ and $x_m$ increases. Moreover, since the neural tangent kernel depends on the neural architecture, this kernel interpretation also suggests that feature extractors of different architectures, including different layers, can have different properties and measure different aspects of the similarity between $x$ and $x_m$. We can see this more clearly later in the investigation of FSSD in different layers.


\section{Our Algorithm}

Based on this phenomenon, we can now construct our OoD detection algorithm. 
Since the uniform noise input can be considered to possess the highest degree of OoD, we use the center of their features as the FSS $F^{\ast}$. The FSSD can then be calculated using Equation~\eqref{fssd}. Note this calculation of FSS $F^{\ast}$ is independent from the choice of in-distribution and OoD datasets. When such natural choice of uniform noise is unavailable, we can choose FSS $F^{\ast}$ to be the center of features of OoD validation data instead.

Since a single forward-pass computation through the network  can give us features from each layer, it is also convenient to calculate FSSDs from different layers and ensemble them as $ \text{FSSD-Ensem} \rbr{x} = \sum_{ k = 1 }^K \alpha_k \text{FSSD}^{\rbr{k}} \rbr{x} $.
The ensemble weights $\alpha_k$ can be determined using logistic regression on some validation data as in~\cite{mahalanobis} (see Evaluation section in Experiments).
In later experiments, if not specified, we use the ensembled FSSD from all layers.
We note that it is also possible to ensemble FSSDs from different architectures or multiple training snapshots~\cite{hor_and_ver_ensemble, snapshot}. This may further enhance the performance of OoD detection.
We investigate the effect of ensembling in the next section. 

Beside, we also  adopt input pre-processing as in ~\cite{odin, mahalanobis} .
The idea is to add small perturbations to the test samples in order to increase the in-distribution score.
It is shown in~\cite{odin, kamoi2020mahalanobis} that in-distribution data are more sensitive to such perturbation and it can therefore enlarge the score gap between in-distribution and OoD samples.
In particular, we perturb as $ \tilde{x} = x + \epsilon \mathop{\text{sign}} \rbr{ \nabla_x \text{FSSD} \rbr{x} } $ and take $ \text{FSSD} \rbr{\tilde{x}} $ as the final score.

We present the pseudo-code of computing $\text{FSSD-Ensem} \rbr{x} $ in Algorithm 1.

\begin{algorithm}[h]
\SetAlgoLined
\DontPrintSemicolon
\KwIn{Test samples $ \bm{x} = \cbr{x^{\text{test}}_n}{}_{ n = 1 }^N $, noise samples $\cbr{x^{\text{noise}}_s}{}_{ s = 1 }^S $, ensemble weights $\alpha_k$, perturbation magnitude $\epsilon$, \\ \phantom{\textbf{Input:}\,} feature extractors $\bcbr{\big}{F_{\rbr{k}}}{}_{k=1}^K$}
\For{\text{each feature extractor} $ \bcbr{\big}{F_{\rbr{k}}}{}_{ k = 1 }^K $}
{
1. Estimate FSS $ F^{\ast}_{\rbr{k}} = \sum_{ s = 1 }^S F_{\rbr{k}} \brbr{\big}{x^{\text{noise}}_s} / S $, where $x^{\text{noise}}_s \sim \mathcal{U}[0,1]$, $ s=1,\cdots,S$\;
2. Add perturbation to test sample:    $\tilde{\bm{{x}}} = \bm{x} + \epsilon \mathop{\text{sign}} (\nabla_{\bm{x}} \bnorm{\big}{F_{\rbr{k}} \rbr{\bm{x}} - F^{\ast}_{\rbr{k}} } ) $\;
3. Calculate $ \text{FSSD}^{(k)} \rbr{\bm{x}} = \bnorm{\big}{ F_{\rbr{k}} \rbr{\tilde{\bm{x}}} - F^{\ast}_{\rbr{k}} } $\;
}
\textbf{Return} $\text{\normalfont FSSD-Ensem} \rbr{\bm{x}} =\sum_{ k = 1 }^K \alpha_k \ \text{\normalfont FSSD}^{(k)} \rbr{\bm{x}}$

\label{fss_alg1} 

\caption{Computation of FSSD-Ensem}
\end{algorithm}

\section{Experiments}\label{experiments}

In this section, we investigate the performance of our FSSD algorithm on various OoD detection benchmarks.
\begin{table*}
\centering
\caption{OoD detection benchmarks used in our experiments.}
\resizebox{0.7\linewidth}{!}{%

\begin{tabular}{ccr@{/}lr@{/}lp{0pt}ccc}
\toprule
& \multicolumn{5}{c}{In-distribution} & & \multicolumn{2}{c}{OoD}  & \multirow{3}*{Data type}
\\
\cmidrule(lr){2-6} \cmidrule(lr){8-9}
& \multirow{2}*{Dataset} & \multicolumn{2}{c}{\#Classes} & \multicolumn{2}{c}{\#Samples} & & \multirow{2}*{Dataset} & \#Samples \\
& & (Train&Test) & (Train&Test) & & & (Test) \ \ \ \ \\
\midrule
A & FMNIST & 10&10 & 60k&10k & & MNIST & 10k & Image \\
B & CIFAR10 & 10&10 & 50k&10k & & SVHN & 26k & Image \\
C & ImageNet (dogs) & 50&50 & 50k&10k & & ImageNet (non-dogs) & 10k & Image \\
D & ImageNet & 1000&1000 & 1281.2k&50k & & ImageNet-C & 50k  & Image \\
E & CelebA (not blurry) & 10122&10122 & 153.8k&38.5k & & CelebA (blurry) & 10.3k  & Image \\
F & MS-1M & 64736&16184 & 2923.6k&50k & & IJB-C & 50k & Image \\
G & Genome (before 2011) & 10&10 & 1000k&1000k & & Genome (after 2016) & 6000k  & Sequence \\
\bottomrule
\end{tabular}%
}
\label{bmks}
\end{table*}

\subsection{Setup}

\paragraph{Benchmarks}
To conduct a thorough test of our method, we consider a wide variety of OoD detection benchmarks. In particular, we consider different scales of datasets and different types of data.
We consider different scales of datasets because large scale datasets tend to have more classes which can introduce more ambiguous data. The ambiguous data are of high classification uncertainty, but are not out-of-distribution.
We list the benchmarks in Table~\ref{bmks}. 

We first consider two common benchmarks from previous OoD detection literature~\cite{duq, ren2019likelihood}: \textbf{(A)} FMNIST~\cite{fmnist} vs.\ MNIST~\cite{mnist} and \textbf{(B)} CIFAR10~\cite{cifar10} vs.\ SVHN~\cite{svhn}.
They are known to be challenging for many methods~\cite{ren2019likelihood, generative_know}.
\textbf{(C)} We also construct ImageNet (dogs), a subset of ImageNet ~\cite{imagenet} , as in-distribution data. 
The OoD data are non-dog images from ImageNet.

For large-scale problems, we consider three benchmarks.
\textbf{(D)} We train models on ImageNet and detect corrupted  images from the ImageNet-C dataset~\cite{corruption}.
We test each method on 80 sets of corruptions (16 types and 5 levels).
\textbf{(E)} 
We train models on face images without the ``blurry'' attribute from CelebA~\cite{celeba} and detect face images with the ``blurry'' attribute.
\textbf{(F)} We train models on web images of celebrities from MS-Celeb-1M (MS-1M)~\cite{ms1m} and detect video captures from IJB-C~\cite{ijbc} which in general have lower quality due to pose, illumination, and resolution issues.
We also consider \textbf{(G)} the bacteria genome benchmark introduced by~\cite{ren2019likelihood}, which consists of sequence data.

To train models on  in-distribution datasets, we follow previous works~\cite{mahalanobis} to train LeNet on FMNIST and ResNet with 34 layers on CIFAR10, ImageNet, and ImageNet (dogs).
For two face recognition datasets (CelebA and MS-1M), we train ResNeXt with 50 layers.
For the genome sequence dataset, we use an character embedding layer and two Bidirectional LSTM layers~\cite{bilstm}.

\paragraph{Compared methods}
We compare our method with the following six common methods for OoD detection. \textbf{\textit{Base}}: the baseline method using the maximum softmax probability $ p \rbr{ \hat{y} \mvert x } $~\cite{hendrycks17baseline}.
\textbf{\textit{ODIN}}:  temperature scaling on  logits  and input pre-processing~\cite{odin}.
\textbf{\textit{Maha}}: Mahalanobis distance of the sample feature to the closest class-conditional Gaussian distribution which is estimated from the training data~\cite{mahalanobis}.
In our experiments, we follow ~\cite{mahalanobis} to use both feature (layer) ensemble and input pre-processing. 
\textbf{\textit{DE}}: Deep Ensemble which averages the softmax probability predictions from multiple independently trained classifiers ~\cite{deepensemble}.
In our experiments, we take the average of 5 classifiers by default.
\textbf{\textit{MCD}}: Monte-Carlo Dropout that uses dropout during both training and inference~\cite{mcdropout}.
We follow ~\cite{Ovadia2019CanYT} to dropout convolutional layers.
For OoD detection, we calculate both the mean and the variance of 32 independent predictions and choose the better one to report.
\textbf{\textit{OE}}: Outlier exposure that explicitly enforces uniform probability prediction on an auxiliary dataset of outliers ~\cite{outlierexposure}.
For the choice of auxiliary datsets, we use KMNIST~\cite{kmnist} for benchmark A, CelebA~\cite{celeba} for benchmark C, and ImageNet-1K~\cite{imagenet} for benchmark B, E, F.
We do not evaluate \textit{OE} on the sequence benchmark, since we can not find a reasonable auxiliary dataset.
We remark here that \textit{Base}, \textit{ODIN}, and \textit{FSSD} can be deployed directly with a trained neural network, \textit{MCD} needs a trained neural network with dropout layers, while \textit{DE} needs multiple trained classifiers.
Besides, \textit{Maha} needs to use the training data during OoD detection on test data and \textit{OE} trains a neural network either from scratch or by fine-tuning to utilize the auxiliary dataset. 

\paragraph{Evaluation}
We follow \cite{ren2019likelihood, outlierexposure} to use the following metrics to assess the performance of OoD detection.
\textbf{AUROC}: Area Under the Receiver Operating Characteristic curve.
\textbf{AUPRC}: Area Under the Precision-Recall Curve.
\textbf{FPR80}: False Positive Rate when the true positive rate is 80\%.

For hyper-parameter tuning, we follow ~\cite{mahalanobis, ren2019likelihood, odin} to use a separate validation set, which consists of 1,000 images from each  in-distribution and OoD data pair.
Ensemble weights $\alpha_k$ for FSSD from different layers are extracted from a logistic regression model, which is trained using nested cross validation within the validation set as in~\cite{mahalanobis, ma2018characterizing}.
The same procedure is performed on \textit{Maha} for fair comparison.
The perturbation magnitude $\epsilon$ of input pre-processing for \textit{ODIN}, \textit{Maha}, and \textit{FSSD} is searched from 0 to 0.2 with step size 0.01.
The temperature $T$ of \textit{ODIN} is chosen from 1, 10, 100, and 1000, and the dropout rate of \textit{MCD} is chosen from 0.01, 0.02, 0.05, 0.1, 0.2, 0.3, 0.4, and 0.5.

\subsection{Main results}
\begin{figure}
    \centering
    \includegraphics[width=\linewidth]{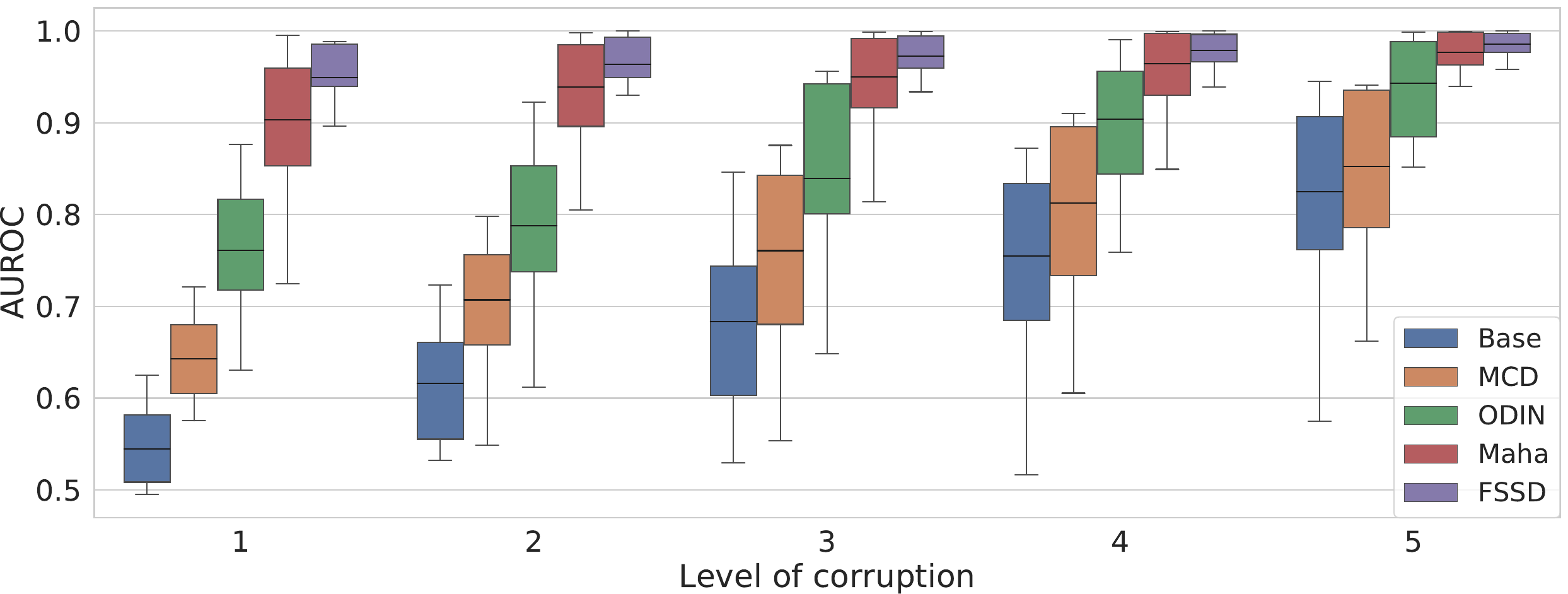}
    \caption{Comparison of AUROC on ImageNet vs.\ ImageNet-C.  FSSD enjoys the highest mean and the least variance across all corruption levels.}
    \label{fig:corruption}
\end{figure}

The main results are presented in Table~\ref{tab:main_results} and Figure~\ref{fig:corruption}. 
In Table~\ref{tab:main_results}, we can see that larger datasets entail greater difficulty in OoD detection.
Notably, the advantage of \textit{FSSD} over other methods increases as the dataset size increases. Other methods like \textit{Maha} and \textit{OE} perform well on some small benchmarks, but have large variance across different datasets.
In comparison, \textit{FSSD} maintains great performance on these benchmarks.
On the genome sequence dataset, we also observe that \textit{FSSD} outperforms other methods.
These results show that \textit{FSSD} is a promising effective method  for a wide range of applications.

Inspired by~\cite{Ovadia2019CanYT}, we also evaluate the methods on the ability of detecting distributional dataset shift like Gaussian noise and JPEG artifacts.
Figure~\ref{fig:corruption} shows the means and quartiles of AUROC of the compared methods over 16 types of corruptions on 5 corruption levels.
We can observe that for each method, the performance of OoD detection increases as the level of corruption increases, while \textit{FSSD} enjoys the highest AUROC and much less variation over different types of corruptions.
The CelebA benchmark also evaluates the methods on detecting the dataset shift of the attribute ``blurry''.
However, all methods including \textit{FSSD} do not perform very well.
There are two possible reasons: (1) the attribute ``blurry'' of CelebA may be annotated not clearly enough; (2) the blurs in the wild may be more difficult to detect than the simulated blurs in ImageNet-C.
Overall, we can see that FSSD can more reliably detect different kinds of distributional shifts.


\begin{table*}[t]
\caption{Main results. All values are in \%.}
\centering
\resizebox{1.8\columnwidth}{!}{%
\begin{tabular}{ccrrrrrrrr}
\toprule
 
                                          &             Datasets (Architecture)       & Metrics & \textit{Base} & \textit{ODIN} & \textit{Maha} & \textit{DE} & \textit{MCD} & \textit{OE} &  \textit{FSSD}  \\
\midrule
\multirow{9}{*}{\makecell{Small-scale \\ benchmarks}}  &                                           
\multirow{3}{*}{\begin{tabular}[c]{@{}c@{}}FMNIST vs.\ MNIST\\  (LeNet)\end{tabular}} & AUROC &  77.3    & 96.9  &  \textbf{99.6}    &  83.9  &  81.7 & \textbf{99.6} & \textbf{99.6}     \\
                      & & AUPRC &  79.2    &  93.0  & \textbf{99.7}  & 83.3 &   85.3 & 99.6  &  \textbf{99.7}    \\
                                    & & FPR80 &  43.5    & 2.5  &  \textbf{0.0}    &  27.5  & 36.8 & \textbf{0.0} & \textbf{0.0}   \\
\cmidrule(l){2-10}
& \multirow{3}{*}{\begin{tabular}[c]{@{}c@{}}CIFAR10 vs.\ SVHN\\  (ResNet34)\end{tabular}} & AUROC & 89.9   & 96.7   &  99.1    &  93.7  &  96.7 & 90.4  &  \textbf{99.5}   \\
                                                                               & & AUPRC &  85.4    & 92.5 & 98.1     &  90.6  &  93.9   & 89.8  & \textbf{99.5}  \\
                                                                               & & FPR80 & 10.1     & 4.7     &   \textbf{0.3}   & 3.7   & 2.4 & 12.5 & 0.4     \\
\cmidrule(l){2-10}
& \multirow{3}{*}{\begin{tabular}[c]{@{}c@{}}ImageNet dogs vs.\ non-dogs\\  (ResNet34)\end{tabular}}                                               & AUROC &  88.5 & 90.8  & 83.3 & 89.0   & 67.2    & 92.5    & \textbf{93.1}  \\
                                                                               &  & AUPRC & 86.1     &  88.6    &  83.0    & 89.0  & 66.9   &  \textbf{92.6}  &  92.5    \\
                                                                               & & FPR80 &   19.5   &  15.2   &   30.1   & 18.8 & 59.2   &   \textbf{7.9}  &  10.2  \\
\midrule
\multirow{6}{*}{\makecell{Large-scale \\ benchmarks}}  &                                      \multirow{3}{*}{ \begin{tabular}[c]{@{}c@{}}CelebA non-blurry vs.\ blurry \\  (ResNeXt50)\end{tabular}}                                               & AUROC &     71.7 & 73.3     &      73.9 &  74.5  & 69.8    &  71.5 & \textbf{78.3}    \\
                                                                               & & AUPRC & 89.9     & 91.4  & 90.9 & 91.4   &  88.7   &  90.7 & \textbf{92.8}   \\
                                                                               & & FPR80 &  52.0    &  50.3     & 46.0 & 47.1   & 53.2    & 54.2 & \textbf{39.2}  \\
\cmidrule(l){2-10}
& \multirow{3}{*}{\begin{tabular}[c]{@{}c@{}}MS-1M vs.\ IJB-C\\  (ResNeXt50)\end{tabular}}                                               & AUROC &   60.0   & 61.3 & 82.5  & 63.0   & 65.5  & 52.6  & \textbf{86.7}  \\
                                                                               & & AUPRC &  53.3    &  55.9     & 80.6     &     56.1   &  59.4 & 46.6  & \textbf{86.1}    \\
                                                                               & & FPR80 &  61.8    &    59.4  &  29.6    &   56.7      & 58.8 & 64.2 & \textbf{22.1}    \\
\midrule
\multirow{3}{*}{\makecell{Sequence \\ benchmark}} & \multirow{3}{*}{ \begin{tabular}[c]{@{}c@{}} Bacteria Genome \\  (LSTM)\end{tabular}}                                               & AUROC &   69.6   &  70.6  &  70.4  & 70.0   & 69.3    &  NA    & \textbf{74.8} \\
                                                                               & & AUPRC &  69.9   & 71.9  &  69.3   &  56.0  & 70.2    & NA  & \textbf{75.8} \\
                                                                               & & FPR80 &  57.4  & 55.9 & 53.7  &  \textbf{30.0} & 58.3 & NA & 47.4   \\
\bottomrule
\end{tabular}%
}
\label{tab:main_results}
\end{table*}

\subsection{Robustness} 

\begin{figure}[t]
\centering
\begin{subfigure}{0.22\textwidth}
  \centering
  \includegraphics[width=1\linewidth]{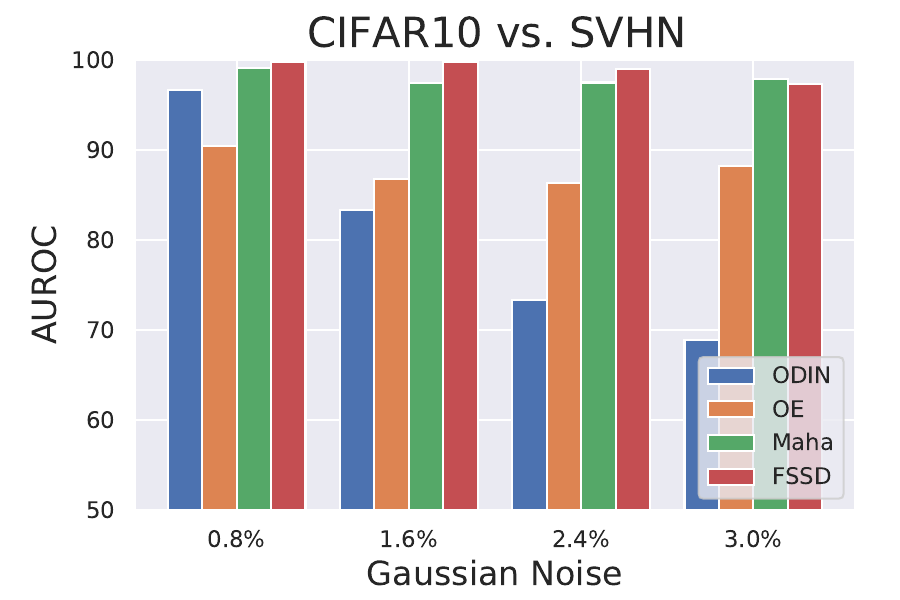} 
\end{subfigure}
\begin{subfigure}{.22\textwidth}
  \centering
  \includegraphics[width=1\linewidth]{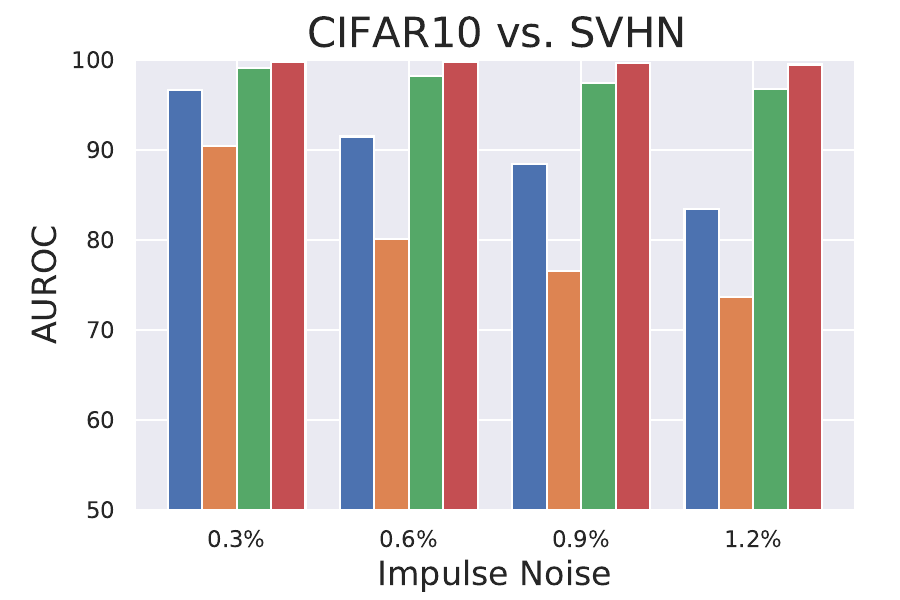}  
\end{subfigure}
\begin{subfigure}{0.22\textwidth}
  \centering
  \includegraphics[width=1\linewidth]{  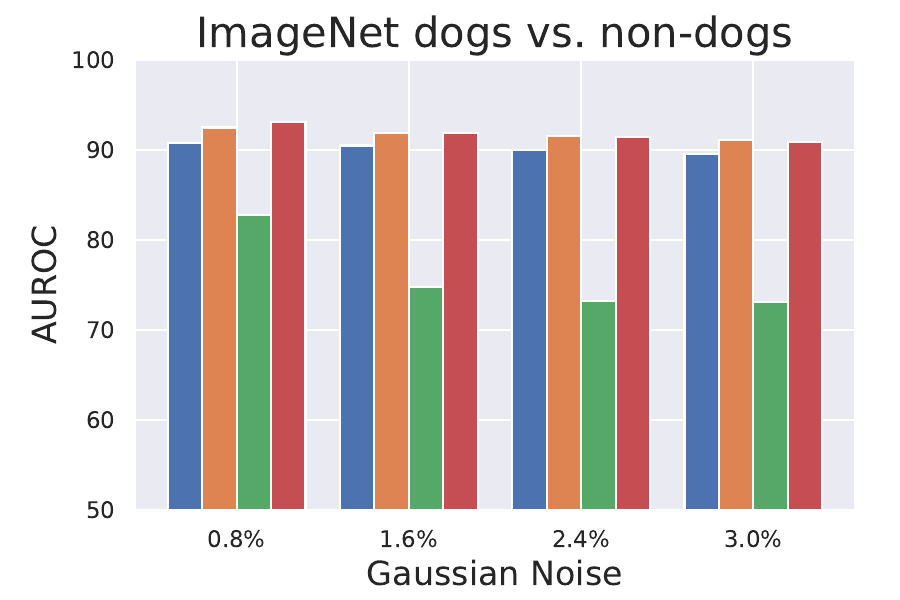} 
\end{subfigure}
\begin{subfigure}{.22\textwidth}
  \centering
  \includegraphics[width=1\linewidth]{ 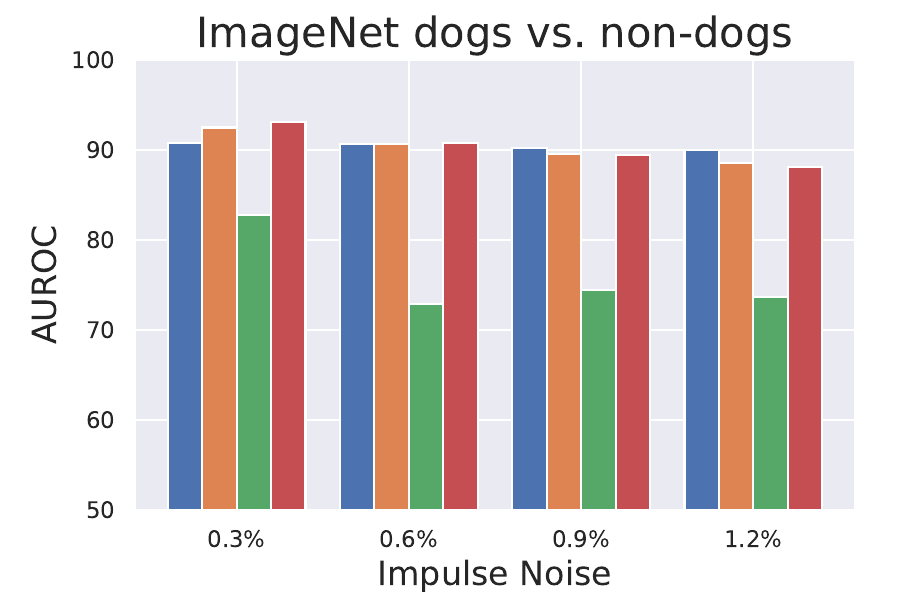}  

\end{subfigure}
\caption{Comparison of OoD detection robustness among methods on slightly corrupted test data.
}
\label{fig:robustness}
\end{figure}

In practice, it is possible that the test data are slightly corrupted or shifted due to the change of data source, e.g., from lab to real world.
We  evaluate the  ability to distinguish in-distribution data from OoD data when test data (both in-distribution and OoD) are slightly corrupted.
Note that we still use non-corrupted data during network training and hyper-parameter tuning.
We apply Gaussian noise and impulse noise, two typical corruptions, with varying levels.
Test results on CIFAR10 vs.\ SVHN and ImageNet dogs vs.\ non-dogs are shown in Figure~\ref{fig:robustness}.
We can see that \textit{FSSD} is robust to corruptions presented in test images, while  other methods may degrade.

\begin{figure*}
\centering
\begin{subfigure}{\textwidth}
  \centering
  \includegraphics[width=1\linewidth]{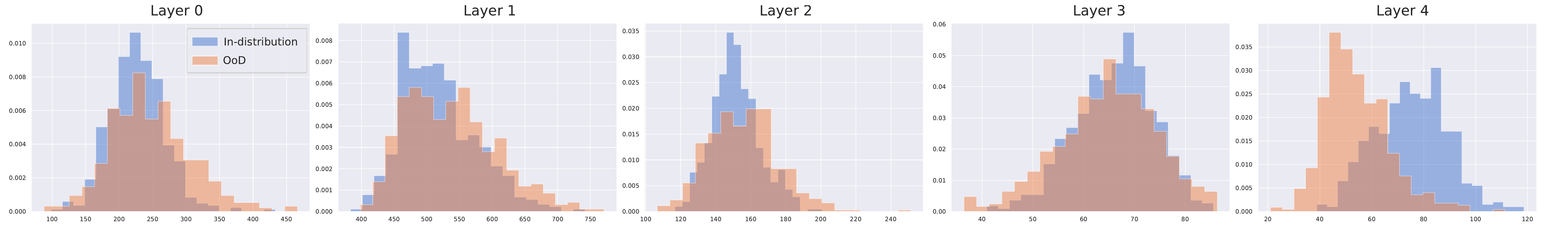}  
  \caption{ImageNet (dogs) vs.\ ImageNet (non-dogs)}
\end{subfigure}
\begin{subfigure}{\textwidth}
  \centering
  \includegraphics[width=1\linewidth]{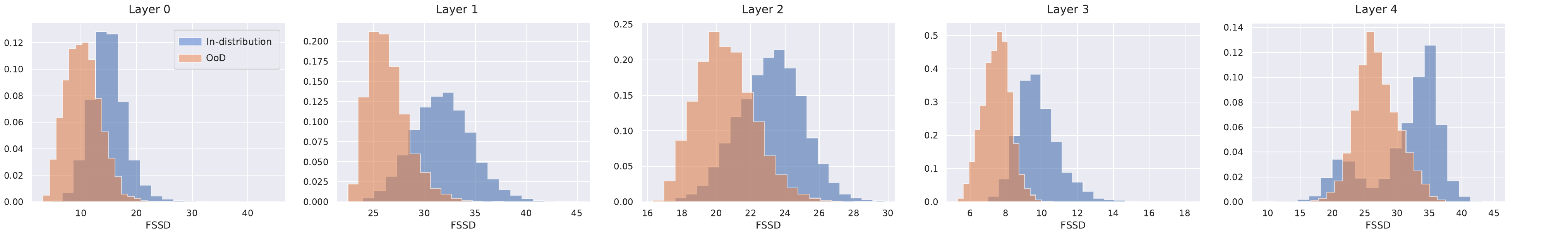}  
  \caption{CIFAR10 vs.\ SVHN}
\end{subfigure}
\caption{FSSDs from different layers behave differently. Each row contains FSSD histograms extracted from different layers of a trained neural network. FSSDs of ImageNet (dogs) and ImageNet (non-dogs) are similar in early layers; while FSSDs of CIFAR10 and SVHN differ in all the layers. This can be explained by the fact that ImageNet (dogs) and ImageNet (non-dogs) are similar in low-level statistics since they are sampled from the same dataset, and that FSSDs in early layers capture more of the difference in low-level statistics.}
\label{fig:layer_ensemble}
\end{figure*}
\begin{figure*}[h]
    \centering
    \begin{subfigure}{0.3\textwidth}
         \centering
         \includegraphics[width=\linewidth]{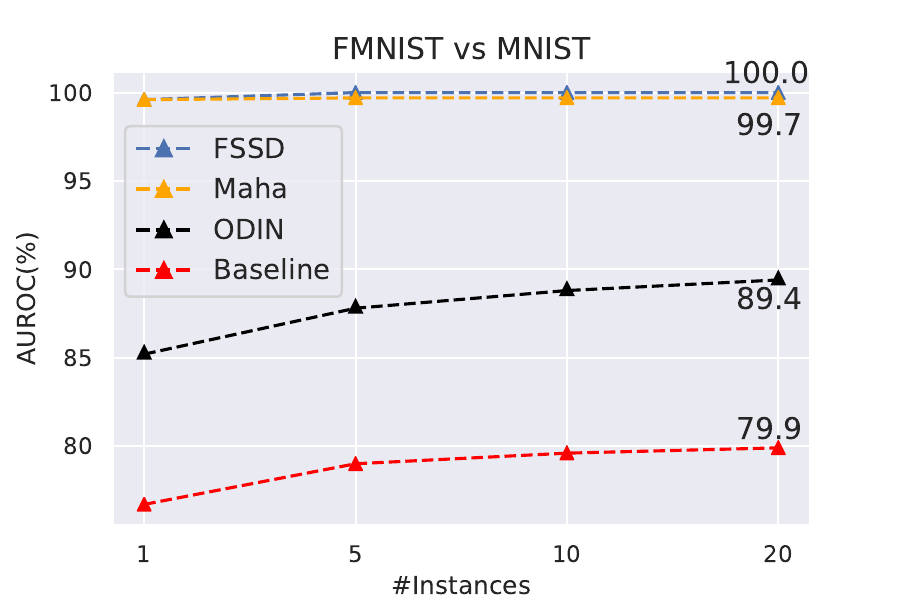}
         \label{dataset2}
    \end{subfigure}
    \begin{subfigure}{.3\textwidth}
         \centering
         \includegraphics[width=\linewidth]{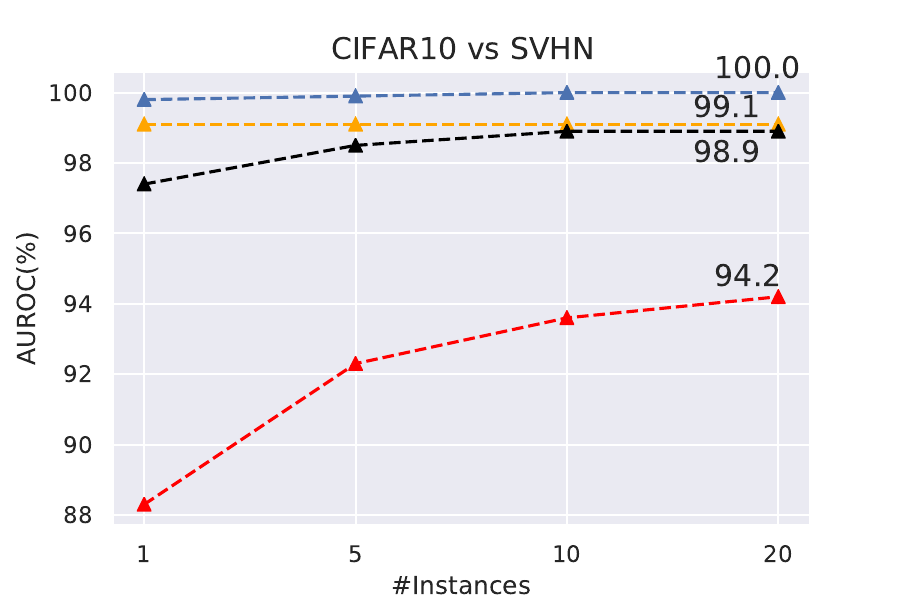}
         \label{dataset2}
    \end{subfigure}
    \begin{subfigure}{.3\textwidth}
         \centering
         \includegraphics[width=\linewidth]{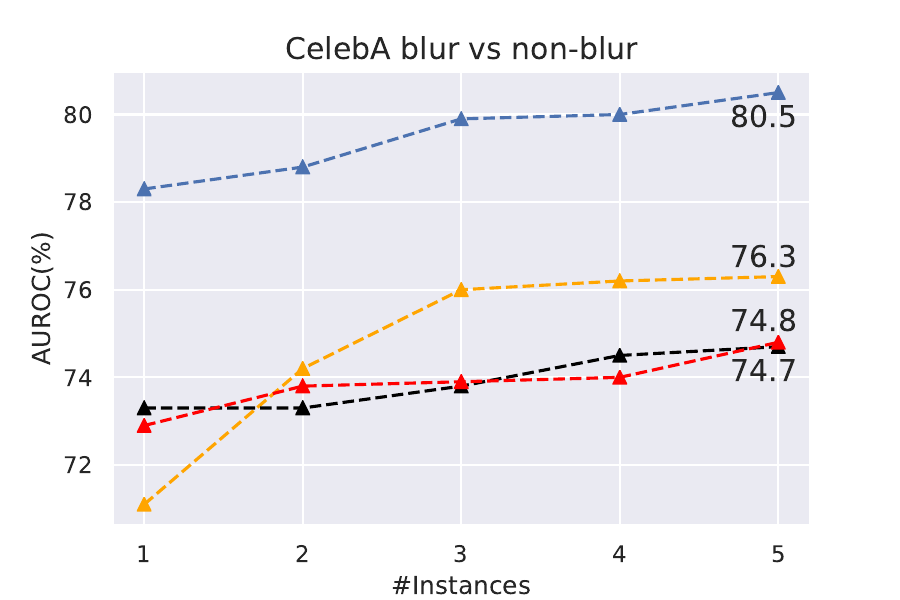}
         \label{dataset2}
    \end{subfigure}
    \caption{Comparison of network ensembles of \textit{Base}, \textit{ODIN}, \textit{Maha}, and \textit{FSSD} scores.
    }
    \label{ensemble}
\end{figure*}

\subsection{Effects of ensemble}

During our experiments, we find that the ensemble plays an important role in enhancing the performance of FSSD. 
Previous studies show that an important issue for ensemble-based algorithms is enforcing diversity~\cite{deepensemble}.
In our case, we find that FSSD has high diversity across different layers, and benefit from such diversity to reach higher performance. In Figure~\ref{fig:layer_ensemble}, we find that FSSD in different layers are working differently. 
This can be explained by previous works on understanding neural networks by visualizing the different representations learned by low and deep layers of a neural network~\cite{vis_eccv2014, scenecnn_iclr15}.
Generally, FSSDs from deep layers reflect more high-level features and FSSDs from early layers reflect more low-level statistics. ImageNet (dogs) and ImageNet (non-dogs) are from the same dataset (ImageNet), and are therefore similar in terms of low-level statistics; while the differences between CIFAR10 and SVHN are in all different levels.
From the perspective of kernel interpretation, this means that the neural tangent kernels of different layers diversify well and allow the ensemble of FSSD to capture different aspects of the discrepancy between the test data and training data.  We show more examples of FSSDs in different layers on our Github page.

Furthermore, Figure~\ref{ensemble}  demonstrates that the performance of FSSD can be further boosted by using network ensembles. Considering the low computational cost of ensembling FSSD, this shows FSSD can be a promising method in scenarios that require high-performance OoD detectors.

\section{Related works}\label{related_work}

\subsection{Out-of-distribution detection}
According to different understandings of OoD samples, previous OoD detection methods can be summarized into four categories.

(1) Some methods regard OoD samples as those with uniform probability prediction across classes~\cite{whyRelu, hendrycks17baseline, odin, confidence-calibrated, adv_manipulation, provably} and treat the test samples with high entropy or low maximum prediction probability as OoD data.
Since these methods are based on prediction, they run the risk of mis-classifying ambiguous data as OoD samples, e.g., when there are thousands of classes in a large-scale dataset.

(2) OoD samples can also be characterized as samples with high epistemic uncertainty which reflects the lack of information on these samples~\cite{deepensemble, mcdropout, WeightUncertainty}.
Specifically, we can propagate the uncertainty of models to the uncertainty of predictions, which characterizes the level of OoD.
\textit{MCD} and \textit{DE} are two popular choices of this type.
However, it is reported that current epistemic uncertainty estimations may noticeably degrade under dataset distributional shift~\cite{Ovadia2019CanYT}.
Our experiments on detecting ImageNet-C from ImageNet (Figure~\ref{fig:corruption}) confirm this.

(3) When the density of data can be approximated, e.g., using generative models~\cite{kingma2018glow, pixelcnn++}, OoD samples can be classified as those with low density.
Recent works provide many inspiring insights on how to improve this idea~\cite{ren2019likelihood, typicality, inputcomplexity}. 
However, these methods typically have extra training difficulty incurred by large generative models.

(4) There are also works designing non-Euclidean metrics to compare test samples to training samples, and regard those with higher distances to training samples as OoD samples~\cite{mahalanobis,duq, kamoi2020mahalanobis, sngp}.
Our approach resembles this type most. Instead of comparing test samples to training samples, we compare the features of the test samples to the center of OoD features.


\subsection{Non-Lipschitz property of neural networks}
Many previous works have reported that the trained neural networks are not Lipschitz continuous~\cite{behrmann_invertible_2019, duq, sngp}. Some OoD detection methods claim that adding Lipschitz constraint may help improve OoD detection performance~\cite{duq, sngp}.
~\cite{bietti_inductive_2019} also demonstrated that the NTK kernel mappings of ReLU networks are  non-Lipschitz. 

Instead of adding Lipschitz constraint, this work utilizes the non-Lipschitz property of neural networks for OoD detection. In fact, the feature space singularity provides us with new understandings of the non-Lipschitz region in the input space, i.e., inputs that are dissimilar to the training data in terms of the NTK.
In the future, it is interesting to further investigate the non-Lipschitz properties of neural networks, e.g., non-Lipschitz behaviours in different layers and how the inductive bias of NTK influences the OoD detection using FSSD.

\section{Conclusion}
In this work, we propose a new OoD detection algorithm based on a novel observation that OoD samples concentrate in the feature space of a trained neural network.
We provide analysis and understanding of the concentration phenomenon by analyzing the training dynamics both theoretically and empirically and further interpreted the algorithm with the neural tangent kernel.
We demonstrate that our algorithm is state-of-the-art in detection performance and is robust to measurement noise. Our further investigation on the effect of ensemble reveals diversity in layer ensembles and shows promising performance of network ensembles.
In summary, we hope that our work can provide new insights for understanding properties of neural networks and add an alternative simple and effective OoD detection method to the safe AI deployment toolkits.

\section*{Acknowledgement}
Bin Dong is supported in part by Beijing Natural Science Foundation (No. 180001); National Natural Science Foundation of China (NSFC) grant No. 11831002 and Beijing Academy of Artificial Intelligence (BAAI).

\bibliography{FSS_bib.bib}

\begin{thebibliography}{47}
\providecommand{\natexlab}[1]{#1}
\providecommand{\url}[1]{\texttt{#1}}
\providecommand{\urlprefix}{URL }
\expandafter\ifx\csname urlstyle\endcsname\relax
  \providecommand{\doi}[1]{doi:\discretionary{}{}{}#1}\else
  \providecommand{\doi}{doi:\discretionary{}{}{}\begingroup
  \urlstyle{rm}\Url}\fi

\bibitem[{Behrmann et~al.(2019)Behrmann, Grathwohl, Chen, Duvenaud, and
  Jacobsen}]{behrmann_invertible_2019}
Behrmann, J.; Grathwohl, W.; Chen, R. T.~Q.; Duvenaud, D.; and Jacobsen, J.-H.
  2019.
\newblock Invertible {Residual} {Networks}.
\newblock In \emph{International {Conference} on {Machine} {Learning}},
  573--582. PMLR.
\newblock \urlprefix\url{http://proceedings.mlr.press/v97/behrmann19a.html}.
\newblock ISSN: 2640-3498.

\bibitem[{Bietti and Mairal(2019)}]{bietti_inductive_2019}
Bietti, A.; and Mairal, J. 2019.
\newblock On the {Inductive} {Bias} of {Neural} {Tangent} {Kernels}.
\newblock \emph{arXiv:1905.12173 [cs, stat]}
  \urlprefix\url{http://arxiv.org/abs/1905.12173}.
\newblock ArXiv: 1905.12173.

\bibitem[{Blundell et~al.(2015)Blundell, Cornebise, Kavukcuoglu, and
  Wierstra}]{WeightUncertainty}
Blundell, C.; Cornebise, J.; Kavukcuoglu, K.; and Wierstra, D. 2015.
\newblock Weight Uncertainty in Neural Network.
\newblock volume~37 of \emph{Proceedings of Machine Learning Research},
  1613--1622. Lille, France: PMLR.
\newblock \urlprefix\url{http://proceedings.mlr.press/v37/blundell15.html}.

\bibitem[{Cao and Gu(2020)}]{ntk2}
Cao, Y.; and Gu, Q. 2020.
\newblock Generalization Error Bounds of Gradient Descent for Learning
  Over-Parameterized Deep ReLU Networks.
\newblock In \emph{The Thirty-Fourth {AAAI} Conference on Artificial
  Intelligence}, 3349--3356. {AAAI} Press.
\newblock
  \urlprefix\url{https://aaai.org/ojs/index.php/AAAI/article/view/5736}.

\bibitem[{Clanuwat et~al.(2018)Clanuwat, Bober-Irizar, Kitamoto, Lamb,
  Yamamoto, and Ha}]{kmnist}
Clanuwat, T.; Bober-Irizar, M.; Kitamoto, A.; Lamb, A.; Yamamoto, K.; and Ha,
  D. 2018.
\newblock Deep Learning for Classical Japanese Literature.

\bibitem[{Gal and Ghahramani(2016)}]{mcdropout}
Gal, Y.; and Ghahramani, Z. 2016.
\newblock Dropout as a Bayesian Approximation: Representing Model Uncertainty
  in Deep Learning.
\newblock volume~48 of \emph{Proceedings of Machine Learning Research},
  1050--1059. New York, New York, USA: PMLR.
\newblock \urlprefix\url{http://proceedings.mlr.press/v48/gal16.html}.

\bibitem[{Guo et~al.(2016)Guo, Zhang, Hu, He, and Gao}]{ms1m}
Guo, Y.; Zhang, L.; Hu, Y.; He, X.; and Gao, J. 2016.
\newblock {MS}-{Celeb}-{1M}: {A} {Dataset} and {Benchmark} for {Large}-{Scale}
  {Face} {Recognition}.
\newblock In Leibe, B.; Matas, J.; Sebe, N.; and Welling, M., eds.,
  \emph{Computer {Vision} – {ECCV} 2016}, volume 9907, 87--102. Cham:
  Springer International Publishing.
\newblock ISBN 978-3-319-46486-2 978-3-319-46487-9.
\newblock \doi{10.1007/978-3-319-46487-9_6}.
\newblock \urlprefix\url{http://link.springer.com/10.1007/978-3-319-46487-9_6}.
\newblock Series Title: Lecture Notes in Computer Science.

\bibitem[{He et~al.(2016)He, Zhang, Ren, and Sun}]{resnet}
He, K.; Zhang, X.; Ren, S.; and Sun, J. 2016.
\newblock Deep {Residual} {Learning} for {Image} {Recognition}.
\newblock In \emph{2016 {IEEE} {Conference} on {Computer} {Vision} and
  {Pattern} {Recognition} ({CVPR})}, 770--778. Las Vegas, NV, USA: IEEE.
\newblock ISBN 978-1-4673-8851-1.
\newblock \doi{10.1109/CVPR.2016.90}.
\newblock \urlprefix\url{http://ieeexplore.ieee.org/document/7780459/}.

\bibitem[{Hein and Andriushchenko(2017)}]{adv_manipulation}
Hein, M.; and Andriushchenko, M. 2017.
\newblock Formal Guarantees on the Robustness of a Classifier against
  Adversarial Manipulation.
\newblock In Guyon, I.; Luxburg, U.~V.; Bengio, S.; Wallach, H.; Fergus, R.;
  Vishwanathan, S.; and Garnett, R., eds., \emph{Advances in Neural Information
  Processing Systems 30}, 2266--2276. Curran Associates, Inc.
\newblock
  \urlprefix\url{http://papers.nips.cc/paper/6821-formal-guarantees-on-the-robustness-of-a-classifier-against-adversarial-manipulation.pdf}.

\bibitem[{Hein, Andriushchenko, and Bitterwolf(2018)}]{whyRelu}
Hein, M.; Andriushchenko, M.; and Bitterwolf, J. 2018.
\newblock Why ReLU Networks Yield High-Confidence Predictions Far Away From the
  Training Data and How to Mitigate the Problem.
\newblock \emph{2019 IEEE/CVF Conference on Computer Vision and Pattern
  Recognition (CVPR)} 41--50.

\bibitem[{Hendrycks and Dietterich(2019)}]{corruption}
Hendrycks, D.; and Dietterich, T. 2019.
\newblock Benchmarking Neural Network Robustness to Common Corruptions and
  Perturbations.
\newblock \emph{Proceedings of the International Conference on Learning
  Representations} .

\bibitem[{Hendrycks and Gimpel(2017)}]{hendrycks17baseline}
Hendrycks, D.; and Gimpel, K. 2017.
\newblock A Baseline for Detecting Misclassified and Out-of-Distribution
  Examples in Neural Networks.
\newblock \emph{Proceedings of International Conference on Learning
  Representations} .

\bibitem[{Hendrycks, Mazeika, and Dietterich(2019)}]{outlierexposure}
Hendrycks, D.; Mazeika, M.; and Dietterich, T. 2019.
\newblock Deep Anomaly Detection with Outlier Exposure.
\newblock In \emph{International Conference on Learning Representations}.
\newblock \urlprefix\url{https://openreview.net/forum?id=HyxCxhRcY7}.

\bibitem[{Hochreiter and Schmidhuber(1997)}]{lstm}
Hochreiter, S.; and Schmidhuber, J. 1997.
\newblock Long Short-Term Memory.
\newblock \emph{Neural Comput.} 9(8): 1735–1780.
\newblock ISSN 0899-7667.
\newblock \doi{10.1162/neco.1997.9.8.1735}.
\newblock \urlprefix\url{https://doi.org/10.1162/neco.1997.9.8.1735}.

\bibitem[{Huang et~al.(2017)Huang, Li, Pleiss, Liu, Hopcroft, and
  Weinberger}]{snapshot}
Huang, G.; Li, Y.; Pleiss, G.; Liu, Z.; Hopcroft, J.~E.; and Weinberger, K.~Q.
  2017.
\newblock Snapshot Ensembles: Train 1, get {M} for free.
\newblock \emph{CoRR} abs/1704.00109.
\newblock \urlprefix\url{http://arxiv.org/abs/1704.00109}.

\bibitem[{Jacot, Gabriel, and Hongler(2018)}]{NTK}
Jacot, A.; Gabriel, F.; and Hongler, C. 2018.
\newblock Neural {Tangent} {Kernel}: {Convergence} and {Generalization} in
  {Neural} {Networks}.
\newblock In Bengio, S.; Wallach, H.; Larochelle, H.; Grauman, K.;
  Cesa-Bianchi, N.; and Garnett, R., eds., \emph{Advances in {Neural}
  {Information} {Processing} {Systems} 31}, 8571--8580. Curran Associates, Inc.
\newblock
  \urlprefix\url{http://papers.nips.cc/paper/8076-neural-tangent-kernel-convergence-and-generalization-in-neural-networks.pdf}.

\bibitem[{Kamoi and Kobayashi(2020)}]{kamoi2020mahalanobis}
Kamoi, R.; and Kobayashi, K. 2020.
\newblock Why is the {Mahalanobis} {Distance} {Effective} for {Anomaly}
  {Detection}?
\newblock \emph{arXiv:2003.00402 [cs, stat]}
  \urlprefix\url{http://arxiv.org/abs/2003.00402}.
\newblock ArXiv: 2003.00402.

\bibitem[{Kingma and Dhariwal(2018)}]{kingma2018glow}
Kingma, D.~P.; and Dhariwal, P. 2018.
\newblock Glow: {Generative} {Flow} with {Invertible} 1x1 {Convolutions}.
\newblock In Bengio, S.; Wallach, H.; Larochelle, H.; Grauman, K.;
  Cesa-Bianchi, N.; and Garnett, R., eds., \emph{Advances in {Neural}
  {Information} {Processing} {Systems} 31}, 10215--10224. Curran Associates,
  Inc.
\newblock
  \urlprefix\url{http://papers.nips.cc/paper/8224-glow-generative-flow-with-invertible-1x1-convolutions.pdf}.

\bibitem[{Krizhevsky(2009)}]{cifar10}
Krizhevsky, A. 2009.
\newblock Learning multiple layers of features from tiny images.
\newblock Technical report.

\bibitem[{Lakshminarayanan, Pritzel, and Blundell(2017)}]{deepensemble}
Lakshminarayanan, B.; Pritzel, A.; and Blundell, C. 2017.
\newblock Simple and Scalable Predictive Uncertainty Estimation using Deep
  Ensembles.
\newblock In Guyon, I.; Luxburg, U.~V.; Bengio, S.; Wallach, H.; Fergus, R.;
  Vishwanathan, S.; and Garnett, R., eds., \emph{Advances in Neural Information
  Processing Systems 30}, 6402--6413. Curran Associates, Inc.
\newblock
  \urlprefix\url{http://papers.nips.cc/paper/7219-simple-and-scalable-predictive-uncertainty-estimation-using-deep-ensembles.pdf}.

\bibitem[{Lakshminarayanan et~al.(2020)Lakshminarayanan, Tran, Liu, Padhy,
  Bedrax-Weiss, and Lin}]{sngp}
Lakshminarayanan, B.; Tran, D.; Liu, J.; Padhy, S.; Bedrax-Weiss, T.; and Lin,
  Z. 2020.
\newblock Simple and Principled Uncertainty Estimation with Deterministic Deep
  Learning via Distance Awareness.
\newblock In \emph{Advances in Neural Information Processing Systems 33}.

\bibitem[{LeCun and Cortes(2010)}]{mnist}
LeCun, Y.; and Cortes, C. 2010.
\newblock {MNIST} handwritten digit database
  \urlprefix\url{http://yann.lecun.com/exdb/mnist/}.

\bibitem[{Lee et~al.(2018{\natexlab{a}})Lee, Lee, Lee, and
  Shin}]{confidence-calibrated}
Lee, K.; Lee, H.; Lee, K.; and Shin, J. 2018{\natexlab{a}}.
\newblock Training Confidence-calibrated Classifiers for Detecting
  Out-of-Distribution Samples.
\newblock In \emph{International Conference on Learning Representations}.
\newblock \urlprefix\url{https://openreview.net/forum?id=ryiAv2xAZ}.

\bibitem[{Lee et~al.(2018{\natexlab{b}})Lee, Lee, Lee, and Shin}]{mahalanobis}
Lee, K.; Lee, K.; Lee, H.; and Shin, J. 2018{\natexlab{b}}.
\newblock A Simple Unified Framework for Detecting Out-of-Distribution Samples
  and Adversarial Attacks.
\newblock In \emph{Proceedings of the 32nd International Conference on Neural
  Information Processing Systems}, NIPS’18, 7167–7177. Red Hook, NY, USA:
  Curran Associates Inc.

\bibitem[{Li, Zhao, and Scheidegger(2020)}]{grandtour}
Li, M.; Zhao, Z.; and Scheidegger, C. 2020.
\newblock Visualizing Neural Networks with the Grand Tour.
\newblock \emph{Distill} \doi{10.23915/distill.00025}.
\newblock Https://distill.pub/2020/grand-tour.

\bibitem[{Li and Liang(2018)}]{ntk1}
Li, Y.; and Liang, Y. 2018.
\newblock Learning Overparameterized Neural Networks via Stochastic Gradient
  Descent on Structured Data.
\newblock In \emph{Proceedings of the 32nd International Conference on Neural
  Information Processing Systems}, NIPS’18, 8168–8177. Red Hook, NY, USA:
  Curran Associates Inc.

\bibitem[{Liang, Li, and Srikant(2018)}]{odin}
Liang, S.; Li, Y.; and Srikant, R. 2018.
\newblock Enhancing The Reliability of Out-of-distribution Image Detection in
  Neural Networks.
\newblock In \emph{International Conference on Learning Representations}.
\newblock \urlprefix\url{https://openreview.net/forum?id=H1VGkIxRZ}.

\bibitem[{Liu et~al.(2015)Liu, Luo, Wang, and Tang}]{celeba}
Liu, Z.; Luo, P.; Wang, X.; and Tang, X. 2015.
\newblock Deep Learning Face Attributes in the Wild.
\newblock In \emph{Proceedings of International Conference on Computer Vision
  (ICCV)}.

\bibitem[{Ma et~al.(2018)Ma, Li, Wang, Erfani, Wijewickrema, Schoenebeck,
  Houle, Song, and Bailey}]{ma2018characterizing}
Ma, X.; Li, B.; Wang, Y.; Erfani, S.~M.; Wijewickrema, S.; Schoenebeck, G.;
  Houle, M.~E.; Song, D.; and Bailey, J. 2018.
\newblock Characterizing Adversarial Subspaces Using Local Intrinsic
  Dimensionality.
\newblock In \emph{International Conference on Learning Representations}.
\newblock \urlprefix\url{https://openreview.net/forum?id=B1gJ1L2aW}.

\bibitem[{Maze et~al.(2018)Maze, Adams, Duncan, Kalka, Miller, Otto, Jain,
  Niggel, Anderson, Cheney, and Grother}]{ijbc}
Maze, B.; Adams, J.; Duncan, J.~A.; Kalka, N.; Miller, T.; Otto, C.; Jain,
  A.~K.; Niggel, W.~T.; Anderson, J.; Cheney, J.; and Grother, P. 2018.
\newblock {IARPA} {Janus} {Benchmark} - {C}: {Face} {Dataset} and {Protocol}.
\newblock In \emph{2018 {International} {Conference} on {Biometrics} ({ICB})},
  158--165. Gold Coast, QLD: IEEE.
\newblock ISBN 978-1-5386-4285-6.
\newblock \doi{10.1109/ICB2018.2018.00033}.
\newblock \urlprefix\url{https://ieeexplore.ieee.org/document/8411217/}.

\bibitem[{Meinke and Hein(2020)}]{provably}
Meinke, A.; and Hein, M. 2020.
\newblock Towards neural networks that provably know when they don't know.
\newblock In \emph{International Conference on Learning Representations}.
\newblock \urlprefix\url{https://openreview.net/forum?id=ByxGkySKwH}.

\bibitem[{Nalisnick et~al.(2019{\natexlab{a}})Nalisnick, Matsukawa, Teh, Gorur,
  and Lakshminarayanan}]{generative_know}
Nalisnick, E.; Matsukawa, A.; Teh, Y.~W.; Gorur, D.; and Lakshminarayanan, B.
  2019{\natexlab{a}}.
\newblock Do Deep Generative Models Know What They Don't Know?
\newblock In \emph{International Conference on Learning Representations}.
\newblock \urlprefix\url{https://openreview.net/forum?id=H1xwNhCcYm}.

\bibitem[{Nalisnick et~al.(2019{\natexlab{b}})Nalisnick, Matsukawa, Teh, and
  Lakshminarayanan}]{typicality}
Nalisnick, E.; Matsukawa, A.; Teh, Y.~W.; and Lakshminarayanan, B.
  2019{\natexlab{b}}.
\newblock Detecting Out-of-Distribution Inputs to Deep Generative Models Using
  Typicality.

\bibitem[{Netzer et~al.(2011)Netzer, Wang, Coates, Bissacco, Wu, and Ng}]{svhn}
Netzer, Y.; Wang, T.; Coates, A.; Bissacco, A.; Wu, B.; and Ng, A.~Y. 2011.
\newblock Reading Digits in Natural Images with Unsupervised Feature Learning.
\newblock In \emph{NIPS Workshop on Deep Learning and Unsupervised Feature
  Learning 2011}.
\newblock
  \urlprefix\url{http://ufldl.stanford.edu/housenumbers/nips2011_housenumbers.pdf}.

\bibitem[{Ovadia et~al.(2019)Ovadia, Fertig, Ren, Nado, Sculley, Nowozin,
  Dillon, Lakshminarayanan, and Snoek}]{Ovadia2019CanYT}
Ovadia, Y.; Fertig, E.; Ren, J.; Nado, Z.; Sculley, D.; Nowozin, S.; Dillon,
  J.~V.; Lakshminarayanan, B.; and Snoek, J. 2019.
\newblock Can You Trust Your Model's Uncertainty? Evaluating Predictive
  Uncertainty Under Dataset Shift.
\newblock In \emph{NeurIPS}.

\bibitem[{Rabanser, Günnemann, and Lipton(2019)}]{FailingLoudly}
Rabanser, S.; Günnemann, S.; and Lipton, Z. 2019.
\newblock Failing {Loudly}: {An} {Empirical} {Study} of {Methods} for
  {Detecting} {Dataset} {Shift}.
\newblock In Wallach, H.; Larochelle, H.; Beygelzimer, A.; Alché-Buc, F.~d.;
  Fox, E.; and Garnett, R., eds., \emph{Advances in {Neural} {Information}
  {Processing} {Systems} 32}, 1396--1408. Curran Associates, Inc.
\newblock
  \urlprefix\url{http://papers.nips.cc/paper/8420-failing-loudly-an-empirical-study-of-methods-for-detecting-dataset-shift.pdf}.

\bibitem[{Ren et~al.(2019)Ren, Liu, Fertig, Snoek, Poplin, Depristo, Dillon,
  and Lakshminarayanan}]{ren2019likelihood}
Ren, J.; Liu, P.~J.; Fertig, E.; Snoek, J.; Poplin, R.; Depristo, M.; Dillon,
  J.; and Lakshminarayanan, B. 2019.
\newblock Likelihood {Ratios} for {Out}-of-{Distribution} {Detection}.
\newblock In Wallach, H.; Larochelle, H.; Beygelzimer, A.; Alché-Buc, F.~d.;
  Fox, E.; and Garnett, R., eds., \emph{Advances in {Neural} {Information}
  {Processing} {Systems} 32}, 14707--14718. Curran Associates, Inc.
\newblock
  \urlprefix\url{http://papers.nips.cc/paper/9611-likelihood-ratios-for-out-of-distribution-detection.pdf}.

\bibitem[{Russakovsky et~al.(2015)Russakovsky, Deng, Su, Krause, Satheesh, Ma,
  Huang, Karpathy, Khosla, Bernstein, Berg, and Fei-Fei}]{imagenet}
Russakovsky, O.; Deng, J.; Su, H.; Krause, J.; Satheesh, S.; Ma, S.; Huang, Z.;
  Karpathy, A.; Khosla, A.; Bernstein, M.; Berg, A.~C.; and Fei-Fei, L. 2015.
\newblock {ImageNet Large Scale Visual Recognition Challenge}.
\newblock \emph{International Journal of Computer Vision (IJCV)} 115(3):
  211--252.
\newblock \doi{10.1007/s11263-015-0816-y}.

\bibitem[{Salimans et~al.(2017)Salimans, Karpathy, Chen, and
  Kingma}]{pixelcnn++}
Salimans, T.; Karpathy, A.; Chen, X.; and Kingma, D.~P. 2017.
\newblock PixelCNN++: A PixelCNN Implementation with Discretized Logistic
  Mixture Likelihood and Other Modifications.
\newblock In \emph{ICLR}.

\bibitem[{Schuster and Paliwal(1997)}]{bilstm}
Schuster, M.; and Paliwal, K. 1997.
\newblock Bidirectional Recurrent Neural Networks.
\newblock \emph{Trans. Sig. Proc.} 45(11): 2673–2681.
\newblock ISSN 1053-587X.
\newblock \doi{10.1109/78.650093}.
\newblock \urlprefix\url{https://doi.org/10.1109/78.650093}.

\bibitem[{Serrà et~al.(2020)Serrà, Álvarez, Gómez, Slizovskaia, Núñez,
  and Luque}]{inputcomplexity}
Serrà, J.; Álvarez, D.; Gómez, V.; Slizovskaia, O.; Núñez, J.~F.; and
  Luque, J. 2020.
\newblock Input Complexity and Out-of-distribution Detection with
  Likelihood-based Generative Models.
\newblock In \emph{International Conference on Learning Representations}.
\newblock \urlprefix\url{https://openreview.net/forum?id=SyxIWpVYvr}.

\bibitem[{van Amersfoort et~al.(2020)van Amersfoort, Smith, Teh, and Gal}]{duq}
van Amersfoort, J.; Smith, L.; Teh, Y.~W.; and Gal, Y. 2020.
\newblock Simple and Scalable Epistemic Uncertainty Estimation Using a Single
  Deep Deterministic Neural Network.

\bibitem[{Xiao, Rasul, and Vollgraf(2017)}]{fmnist}
Xiao, H.; Rasul, K.; and Vollgraf, R. 2017.
\newblock Fashion-MNIST: a Novel Image Dataset for Benchmarking Machine
  Learning Algorithms.

\bibitem[{Xie, Xu, and Zhang(2013)}]{hor_and_ver_ensemble}
Xie, J.; Xu, B.; and Zhang, C. 2013.
\newblock Horizontal and Vertical Ensemble with Deep Representation for
  Classification.
\newblock \emph{CoRR} abs/1306.2759.
\newblock \urlprefix\url{http://arxiv.org/abs/1306.2759}.

\bibitem[{Xie et~al.(2017)Xie, Girshick, Dollar, Tu, and He}]{resnext}
Xie, S.; Girshick, R.; Dollar, P.; Tu, Z.; and He, K. 2017.
\newblock Aggregated {Residual} {Transformations} for {Deep} {Neural}
  {Networks}.
\newblock In \emph{2017 {IEEE} {Conference} on {Computer} {Vision} and
  {Pattern} {Recognition} ({CVPR})}, 5987--5995. Honolulu, HI: IEEE.
\newblock ISBN 978-1-5386-0457-1.
\newblock \doi{10.1109/CVPR.2017.634}.
\newblock \urlprefix\url{http://ieeexplore.ieee.org/document/8100117/}.

\bibitem[{Zeiler and Fergus(2014)}]{vis_eccv2014}
Zeiler, M.; and Fergus, R. 2014.
\newblock Visualizing and understanding convolutional networks.
\newblock In \emph{Computer Vision, ECCV 2014 - 13th European Conference,
  Proceedings}, number PART 1 in Lecture Notes in Computer Science (including
  subseries Lecture Notes in Artificial Intelligence and Lecture Notes in
  Bioinformatics), 818--833. Springer Verlag.
\newblock ISBN 9783319105895.
\newblock \doi{10.1007/978-3-319-10590-1_53}.
\newblock 13th European Conference on Computer Vision, ECCV 2014 ; Conference
  date: 06-09-2014 Through 12-09-2014.

\bibitem[{Zhou et~al.(2015)Zhou, Khosla, Lapedriza, Oliva, and
  Torralba}]{scenecnn_iclr15}
Zhou, B.; Khosla, A.; Lapedriza, A.; Oliva, A.; and Torralba, A. 2015.
\newblock Object Detectors Emerge in Deep Scene CNNs.
\newblock In \emph{International Conference on Learning Representations
  (ICLR)}.

\end{thebibliography}

\end{document}